\documentclass[runningheads]{llncs}

\usepackage{eccv}

\usepackage{colortbl}
\usepackage{todonotes}
\usepackage{pifont}
\usepackage{tikz}
\usepackage{adjustbox}
\usepackage{multirow}
\usepackage{makecell}
\usepackage{booktabs,array}
\usetikzlibrary{shapes,arrows,positioning}

\usepackage{xcolor}
\usepackage{graphicx}
\usepackage{rotating}
\usepackage{tabularx}
\usepackage{amssymb}
\usepackage{pifont}
\usepackage{wrapfig}
\usepackage{microtype}
\newcommand{\xmark}{\ding{55}}%
\usepackage{booktabs}
\usepackage{placeins}

\newcommand{\greencheck}{{\color{green}\checkmark}}
\newcommand{\redxmark}{{\color{red}\xmark}}

\usepackage{eccvabbrv}

\usepackage{graphicx}
\usepackage{booktabs}
\usepackage{marvosym}

\usepackage[accsupp]{axessibility}  

\usepackage{xcolor}

\usepackage[
    colorlinks=true,
    linkcolor=blue,
    citecolor=blue,
    urlcolor=blue
]{hyperref}

\usepackage{orcidlink}

\begin{document}

\title{SemCityLoc: Aerial 6DoF Localization Using Semantic 3D City Models} 

\titlerunning{SemCityLoc: Aerial 6DoF Localization Using Semantic 3D City Models}

\author{Jingfeng Mao\inst{1,2}\orcidlink{0009-0004-0941-1452} \and
Xuyang Chen\inst{1}\orcidlink{0009-0004-4154-3463} \and
Qilin Zhang\inst{1,2,4,5}\orcidlink{0009-0000-9064-1976} \and
Oussema Dhaouadi\inst{1,2,4,6}\orcidlink{0009-0008-6842-5220} \and
Guangming Wang\inst{2}\orcidlink{0000-0002-7675-543X} \and
Brian Sheil\inst{2}\orcidlink{0000-0002-1462-1401} \and
Daniel Cremers\inst{1,4}\orcidlink{0000-0002-3079-7984} \and \\
Yan Xia\inst{3}\orcidlink{0000-0001-6684-9814} \and 
Olaf Wysocki\inst{2}\orcidlink{0000-0002-0016-0229}}
\authorrunning{J.~Mao et al.}

\institute{
Technical University of Munich, Munich, Germany
\and
CV4DT, University of Cambridge, Cambridge, United Kingdom
\and
University of Science and Technology of China, Hefei, China
\and
Munich Center for Machine Learning, Munich, Germany
\and 
Karlsruhe Institute of Technology, Karlsruhe, Germany
\and
ETH Zurich, Zurich, Switzerland
\\
\email{jingfeng.mao@tum.de} 
\email{okw24@cam.ac.uk}}

\maketitle
\begin{figure}[h]
    \centering
    \vspace{-5.5mm}
    \includegraphics[width=0.45\linewidth]{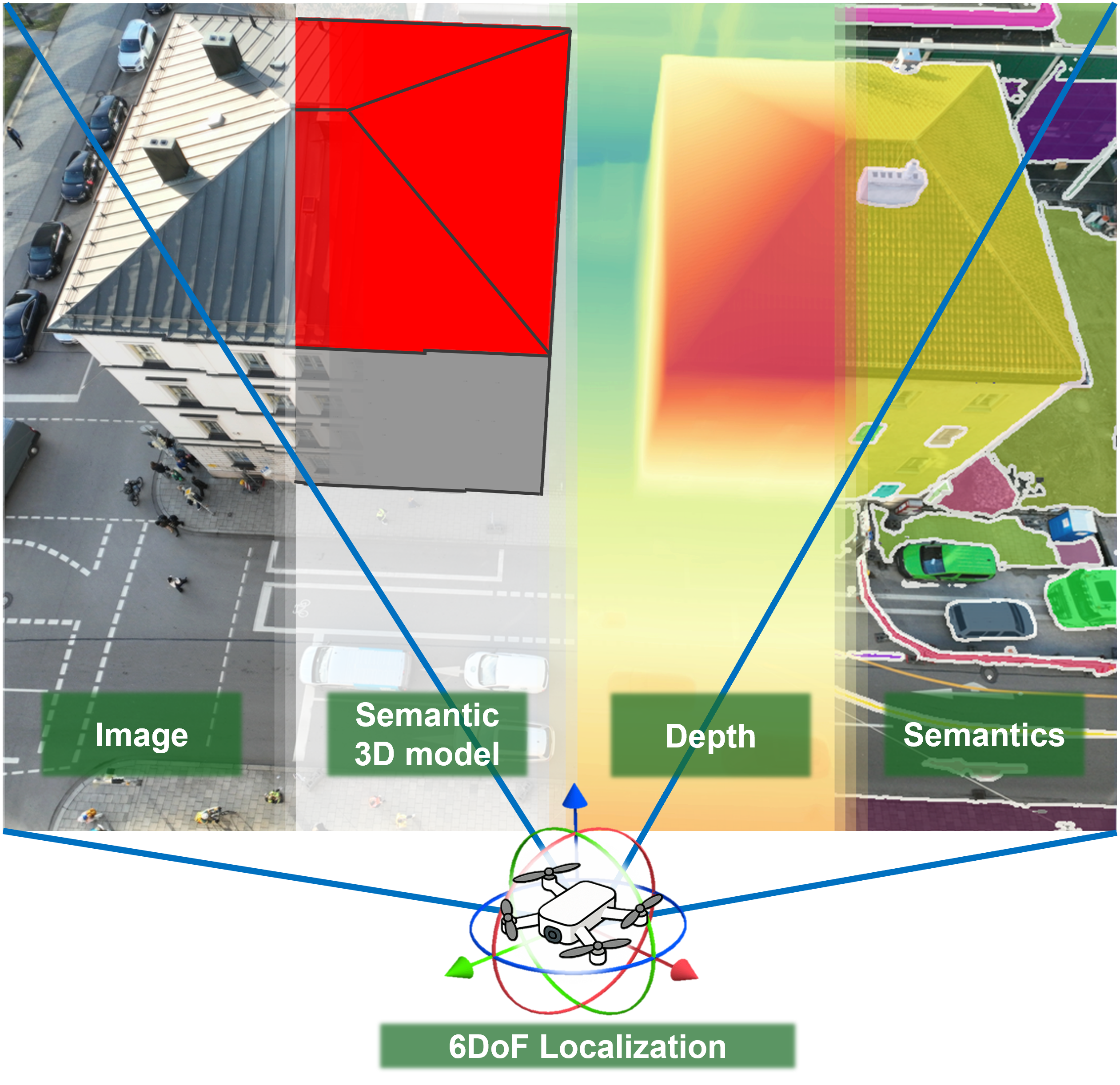}
    \caption{\textbf{SemCityLoc.} We estimate 6DoF UAV poses by aligning foundation-model semantics and depth with lightweight, semantically structured 3D city models—without radiometric scene reconstructions. We further introduce \textbf{SemCityLockeD}, a centimeter-accurate benchmark combining standardized LoD city models with challenging low-altitude UAV imagery.}
    \label{fig:6dofteaser}
    \vspace{-1\baselineskip}
\end{figure}
\begin{abstract}
Aerial 6DoF localization typically relies on precise GNSS signals or radiometrically rich 3D reconstructions, limiting scalability and on-board deployment. 
We propose \textbf{SemCityLoc}, a semantic–geometric alignment system that reframes aerial pose estimation as structured surface registration between foundation-model-derived visual priors and standardized LoD-compliant 3D city models. 
Instead of matching sparse contours or dense texture, our method aligns semantic surfaces and monocular depth with lightweight semantic 3D building models, increasing pose discriminability in repetitive and occluded urban environments. 
To enable accurate evaluation, we introduce \textbf{SemCityLockeD}, the first real-world benchmark combining centimeter-accurate UAV poses with standardized LoD1–LoD3 semantic city models and challenging low-altitude imagery. 
Experiments demonstrate substantial improvements over existing map-based approaches, improving recall by up to 36\% and reducing mean positional error from 9.89\,m to 2.62\,m in challenging urban canyons. 
Our results indicate that semantically structured geometry provides sufficient and scalable constraints for high-precision aerial localization without radiometric scene reconstructions.
The code and data are available at \url{https://albertchen98.github.io/SemCityLoc}.
\end{abstract}    
\section{Introduction}
\label{sec:intro}
Aerial localization for uncrewed aerial vehicles (UAVs) is essential for urban navigation, inspection, mapping, and digital twinning~\cite{tum2twin,dhaouadi2025ortholoc,zhang2025gs4buildings,gaisbauer2025glue,nex2014uav}. Accurate six degrees-of-freedom (6DoF) pose estimation enables operation in global navigation satellite systems (GNSS)-denied environments and supports downstream tasks such as reconstruction and change detection \cite{yushengChangeDetectionReview,wysocki2023scan2lod3}. Compared to ground-level localization, aerial imagery exhibits larger baselines, reduced structural detail, and dominant vanishing directions, resulting in weaker geometric constraints~\cite{gaisbauer2025glue,taira2018inloc,sarlin2020superglue,xia2025trafficloc,bieringer2024analyzing,yan2023render}.

Conventional aerial localization approaches combine real-time kinematic (RTK)-supported sensing with dense textured 3D meshes~\cite{nex2014uav,nex2024usegeo}.
While accurate, such radiometrically rich models are memory-intensive, computationally demanding, and difficult to scale, and may raise privacy concerns in urban environments~\cite{ye2025exploring,yan2023render,zhu2024lod}. Lightweight alternatives are therefore desirable.

Recent work has explored reducing reliance on dense radiometry. OrthoLoC~\cite{dhaouadi2025ortholoc} aligns UAV imagery with orthophotos or digital surface models (DSMs) but remains 2D texture-dependent. LoD-Loc v1 and v2~\cite{zhu2024lod,zhu2025lod} demonstrate localization using abstract building models, yet rely primarily on contour-based alignment and scene-specific training, limiting robustness in close-range urban scenarios~\cite{zhu2024lod,zhu2025lod}. 

Progress is further constrained by benchmark limitations: Many datasets focus on high-altitude flights, provide inconsistent pose accuracy, partially rely on synthetic imagery, or lack standardized level of detail (LoD) definitions despite established CityGML standards~\cite{yan2022crossloc,li2023matrixcity,wu2024uavd4l,ji2025game4loc,dhaouadi2025ortholoc,kolbeOGCCityGeography2021,kolbe2008citygml}. As such, systematic evaluation in realistic low-altitude urban canyon settings remains limited.

We argue that semantically structured 3D city models provide sufficient and scalable constraints for high-precision aerial localization when combined with modern vision priors. We present \textbf{SemCityLoc}, a semantic–geometric alignment system for aerial 6DoF pose estimation. Our method leverages foundation-model-based semantic segmentation and monocular depth and aligns these cues with lightweight, standardized semantic 3D city models (\cref{fig:6dofteaser}). By integrating semantic surface classes with geometric structure, SemCityLoc avoids dense radiometric reconstructions while maintaining strong pose observability, enabling scalable and privacy-preserving localization in GNSS-denied urban environments.

To enable rigorous evaluation in this setting, we introduce \textbf{SemCityLockeD}, a real-world benchmark pairing centimeter-accurate UAV poses with standardized LoD semantic 3D city models and challenging low-altitude imagery. Unlike prior datasets, it enables accurate evaluation across LoD-compliant definitions and close-range urban canyon scenarios.
Our contributions are as follows:
\begin{itemize}
\item \textbf{Semantic–geometric aerial localization.} We propose SemCityLoc, an approach that reformulates aerial localization as structured semantic–geometric surface alignment between foundation-model predictions and lightweight LoD city models.
\item \textbf{Real-world benchmark.} We present SemCityLockeD, a challenging dataset combining centimeter-accurate UAV poses, close-range urban imagery, and standardized LoD1–LoD3 semantic city models for accurate evaluation of map-based aerial localization.
\item \textbf{Extensive evaluation.} We demonstrate consistent improvements over state-of-the-art map-based localization approaches across our and other benchmarks and challenging urban scenarios.
\end{itemize}

\section{Related Works}
\label{sec:rw}
\begin{figure}[htb]
    \centering
    \includegraphics[width=0.5\linewidth]{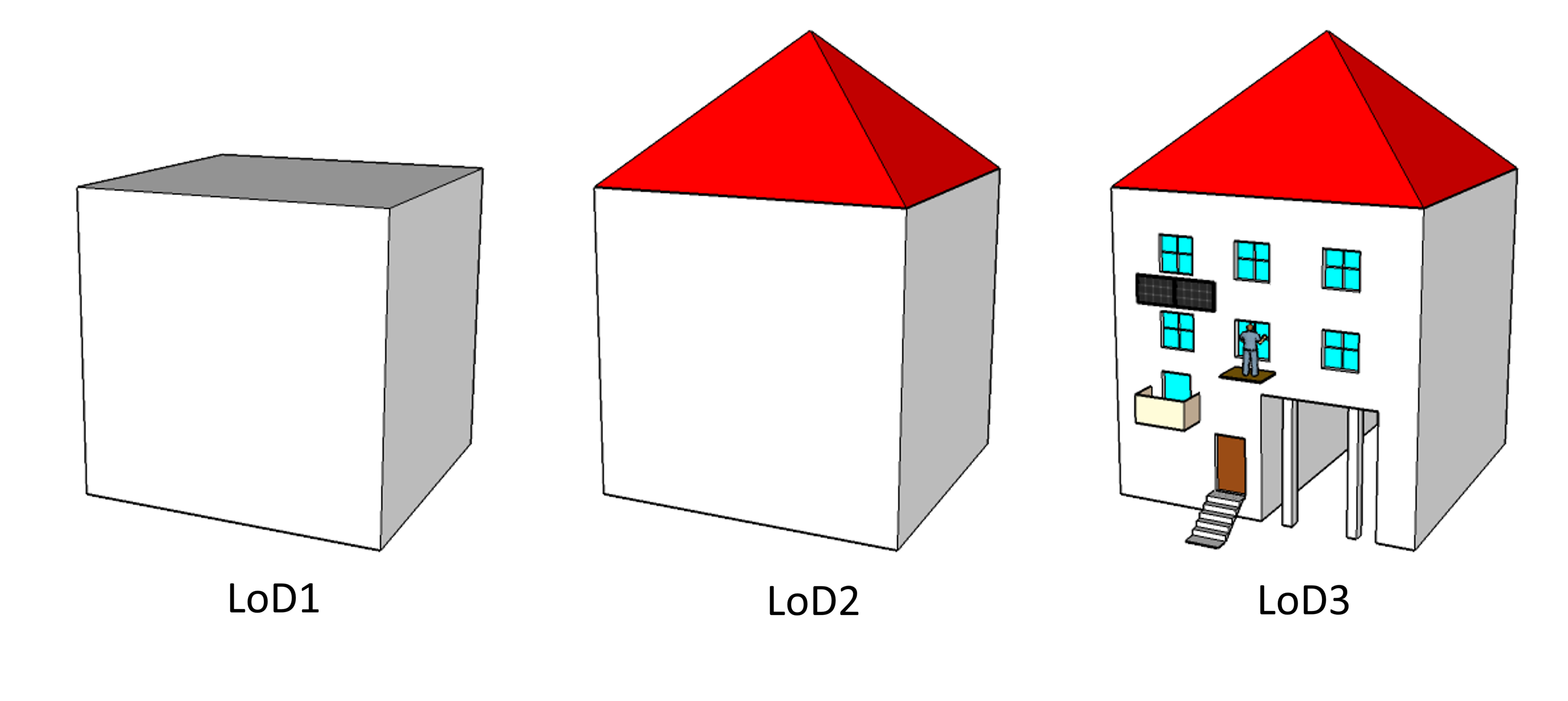}
\caption{\textbf{CityGML Levels of Detail (LoDs).} 
Hierarchical semantic 3D building representations following the CityGML standard \cite{biljecki2014formalisation,citygml2objv2,kolbeOGCCityGeography2021,wysocki2024reviewing}. 
LoD1 models simplified building volumes, LoD2 adds detailed roof structures, and LoD3 further incorporates facade elements. 
Surfaces are semantically classified, enabling structured semantic–geometric reasoning.}
    \vspace{-1\baselineskip}
    \label{fig:lods}
\end{figure}

\noindent \textbf{Standardized LoD Definitions and Semantic 3D City Models.}
Level-of-Detail (LoD) definitions in semantic 3D city models follow the international CityGML 3.0 standard \cite{kolbeOGCCityGeography2021,citygml2objv2,biljecki2014formalisation}, illustrated in \Cref{fig:lods}. 
CityGML encodes hierarchical geometric and semantic information using surface boundary representations, where surfaces (e.g., roof, wall) are explicitly classified. 
LoD1–LoD3 progressively refine structural detail while preserving standardized georeferencing in global coordinate systems.
Governmental CityGML datasets adhere to strict quality standards and often provide centimeter-level geometric accuracy \cite{RoschlaubBatscheider,chen2025public}. 
Recent surveys report more than 216 million openly available semantic 3D building models worldwide \cite{wysocki2024reviewing,awesomeCityGML}. 
Their standardized semantics, structured geometry, and global georeferencing make them a scalable and privacy-preserving alternative to radiometric scene reconstructions.

\noindent \textbf{Aerial Localization Benchmarks.}
Existing UAV localization benchmarks are limited in combining real-world imagery, accurate ground-truth poses, semantic 3D models, and standardized LoD definitions (\cref{tab:uav_datasets}). 
Many rely partially or entirely on synthetic data, lack centimeter-accurate poses, or focus on higher-altitude acquisitions (more than 75m).
\begin{table*}[!htb]
    \captionsetup{size=footnotesize}
    \centering
    \caption{\textbf{Comparison of urban UAV benchmark datasets for 6DoF aerial localization.} Our SemCityLockeD dataset is the first to combine real-world imagery, accurate RTK-supported poses, close-range challenges, semantic 3D city models, and standardized LoD representations in a single benchmark. \greencheck = available/supported; \redxmark = not available; $\thicksim$ = partially available.}
    \label{tab:uav_datasets}
    \begin{adjustbox}{width=1\textwidth, center}
        \begin{tabular}{lcccccccc}
            \toprule
            Name  & Year & View & Real world? & Accurate poses? & Close-range challenge? & Semantic 3D city models? & Standardized LoD? \\
              &  &  &  & (GCP/RTK-supported) &  &  &  \\
            \midrule
            CrossLoc \cite{yan2022crossloc} & 2022 & oblique + nadir & mixed & \redxmark & \redxmark & \redxmark & \redxmark  \\
            MatrixCity \cite{li2023matrixcity} & 2023 & oblique & synthetic & $\thicksim$ & $\thicksim$ & \redxmark & \redxmark  \\
            UAVD4L \cite{wu2024uavd4l} & 2024 & oblique + nadir & mixed & \redxmark & \redxmark & \redxmark & \redxmark  \\
            Swiss-EPFL \cite{zhu2024lod} & 2024 & oblique + nadir & mixed & \redxmark & \redxmark & \greencheck & \redxmark \\
            UAVD4L-LoD \cite{zhu2024lod} & 2024 & oblique + nadir & mixed & \redxmark & \redxmark & \greencheck & \redxmark \\
            GTA-UAV \cite{ji2025game4loc} & 2025 & nadir & synthetic & $\thicksim$ & $\thicksim$ & \redxmark & \redxmark \\ 
            OrthoLoC \cite{dhaouadi2025ortholoc} & 2025 & oblique + nadir & real & \greencheck & $\thicksim$ & \redxmark & \redxmark  \\ \hline
            \textbf{SemCityLockeD (ours)} & 2025 & oblique + nadir & real & \greencheck & \greencheck  & \greencheck & \greencheck \\
            \bottomrule
        \end{tabular}
    \end{adjustbox}
\end{table*}
OrthoLoC~\cite{dhaouadi2025ortholoc} provides real-world imagery with RTK-supported poses but does not include semantic 3D models. 
Swiss-EPFL and UAVD4L-LoD~\cite{zhu2024lod} introduce image–LoD pairs, yet their LoD representations do not strictly follow established CityGML standards and pose inconsistencies have been reported in prior works \cite{dhaouadi2025ortholoc}.
While updated versions have been announced \cite{zhu2025lod}, they are not publicly available at the time of writing.
Furthermore, most datasets underrepresent close-range urban canyon environments.
In contrast, our SemCityLockeD provides real-only low-altitude urban canyon scenarios with cm-grade standard-compliant 3D LoD models and pose pairings, enabling accurate testing in challenging scenarios.

\noindent \textbf{Map-Based and Semantic Aerial Localization.}
Classical visual localization relies on dense SfM reconstructions~\cite{schonberger2016structure}, followed by feature matching and PnP-based pose estimation~\cite{sarlin2020superglue,barath2019magsac}. 
While accurate, these approaches depend on radiometrically rich and memory-intensive scene models that are difficult to scale in urban environments.
Reconstruction-free methods leverage publicly available map data. 
OrienterNet~\cite{sarlin2023orienternet} and MapLocNet~\cite{wu2024maplocnet} use navigation maps but are limited to 3-DoF. 
OrthoLoC~\cite{dhaouadi2025ortholoc} achieves 6-DoF via cross-modal matching with orthophotos and DSMs.
On the other hand, lightweight LoD models provide a promising alternative. 
LoD-Loc~\cite{zhu2024lod} aligns neural wireframes with projected building edges, and LoD-Loc v2~\cite{zhu2025lod} extends this via silhouette alignment. 
However, these approaches rely primarily on contour cues without explicitly incorporating surface semantics or depth geometry.
In contrast, we jointly leverage foundation models of semantic segmentation and monocular depth within standardized semantic 3D city models, enabling robust semantic–geometric alignment beyond contour-based matching.
Recent foundation models have significantly improved segmentation~\cite{simeoni2025dinov3} and monocular depth estimation~\cite{wang2025moge}. 
In our framework, these components serve as modular perceptual priors for the localization.

\section{Method}
\begin{figure*}[htb]
    \centering
    \includegraphics[width=1.0\linewidth]{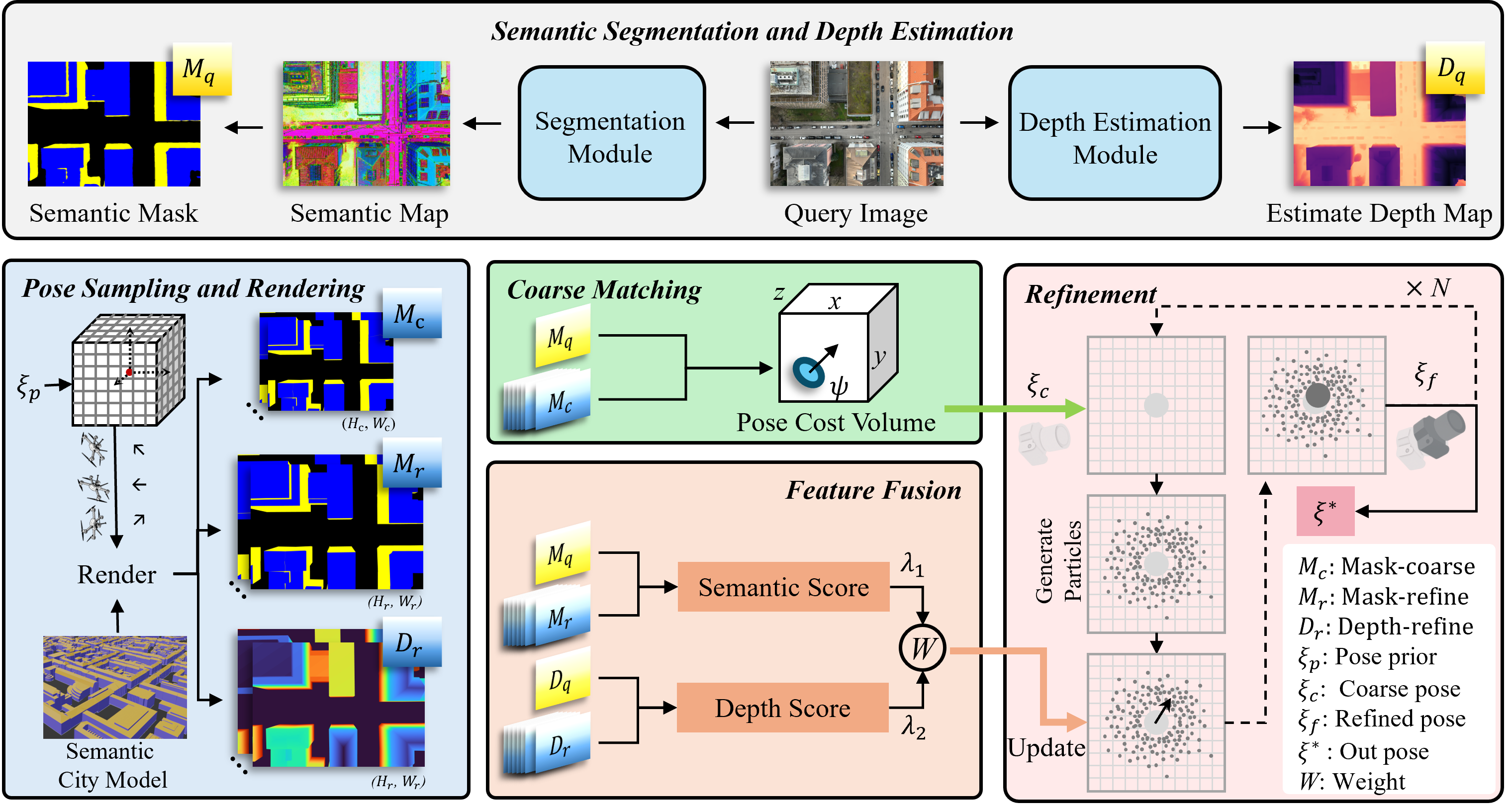}
    \caption{\textbf{Overview of SemCityLoc.} SemCityLoc estimates the 6DoF camera pose $\boldsymbol{\xi}^*$ of a UAV query image via coarse-to-fine semantic--geometric alignment against a semantic LoD city model. \textbf{Inputs} are the query image $I_q$, a prior pose $\boldsymbol{\xi}_p$, and the semantic 3D city model.
    \emph{(Top) Semantic Segmentation and Depth Estimation:} a DINOv3 backbone with a lightweight DPT decoder predicts the query semantic mask $M_q$, while MoGe-2 predicts the query depth map $D_q$.
    \emph{(Bottom-left) Pose Sampling and Rendering:} we uniformly sample candidate poses around the prior $\boldsymbol{\xi}_p$ and render, from the city model, the coarse semantic mask $M_c$, the fine semantic mask $M_r$, and the depth map $D_r$ (coarse masks at resolution $(H_c,W_c)$, fine masks at twice the resolution $(H_r,W_r)$).
    \emph{(Coarse Matching):} comparing each rendered mask $M_c$ with $M_q$ through a per-class IoU semantic cost $C^s$ builds a 4D cost volume over $(x,y,z,\psi)$, whose maximum yields the coarse pose $\boldsymbol{\xi}_c$.
    \emph{(Refinement):} starting from $\boldsymbol{\xi}_c$, a particle filter generates perturbed pose hypotheses; each is scored by \emph{Feature Fusion}, which combines a semantic score ($M_q$ vs.\ $M_r$, cost $C^s$) and a depth score ($D_q$ vs.\ $D_r$, cost $C^d$) into a single weight $W=\lambda_1 C^s+\lambda_2 C^d$. Particles are updated in a weighted manner, and the process repeats for $N$ iterations, refining $\boldsymbol{\xi}_c \!\rightarrow\! \boldsymbol{\xi}_f$ and producing the final pose $\boldsymbol{\xi}^*$.} 
    \label{fig:method}
    \vspace{-1\baselineskip}
\end{figure*}

As illustrated in \Cref{fig:method}, SemCityLoc follows a coarse-to-fine semantic–geometric alignment strategy. Given a query image and an initial pose prior, we first perform a 4D semantic cost-volume search to obtain a coarse pose estimate. We then refine this estimate via joint semantic and depth alignment using a particle-filter optimization scheme to recover the final 6DoF camera pose.
\noindent\textbf{Problem Reformulation.} 
Our goal is to estimate the 6DoF camera pose 
$\boldsymbol{\xi}^* = (x, y, z, \phi, \theta, \psi) $ 
where $(x,y,z)$ denote translation and $(\phi,\theta,\psi)$ denote roll, pitch, and yaw. Since roll and pitch can be reliably estimated from gravity measurements, we keep $(\phi,\theta)$ fixed and perform search only over $(x,y,z,\psi)$.
The problem is formulated as follows (\cref{fig:method}): Given a semantic LoD city model $\mathcal{M}$ of a target urban area, a UAV-acquired image $I$, and an initial coarse camera pose $\boldsymbol{\xi}_{pri}$, the proposed method aims to determine the accurate 6DoF camera pose $\boldsymbol{\xi}^*$ by jointly leveraging the semantic and geometric features extracted from the image and the 3D semantic city model. 

Unlike contour-based alignment, which relies on sparse edge correspondences, semantic surface partitions introduce region-level structural constraints that increase pose observability. 
By aligning semantically labeled roof and facade surfaces rather than isolated edges, the search space is constrained by both geometric consistency and class-specific spatial arrangement. 
This reduces ambiguity in repetitive architectural patterns and improves robustness under occlusion and oblique viewing conditions. 
In essence, semantic alignment transforms the localization problem from edge matching to structured surface registration.

\subsection{Coarse Pose Selection}
\label{Sec:Coarse}
\textbf{Semantic Segmentation.} 
DINOv3 represents a recently introduced vision foundation model that learns powerful and generalizable global and local representations via self-supervised learning~\cite{simeoni2025dinov3}. 
In semantic segmentation, remarkable performance can be achieved by freezing the pretrained backbone and training a lightweight decoder tailored for the task~\cite{yang2025segdino,huang2025real}. 
Specifically, we employ a pretrained DINOv3 Vision Transformer~\cite{simeoni2025dinov3} as a backbone together with a lightweight DPT decoder~\cite{ranftl2021vision} to predict semantic segmentation masks for the query image $I_q$:
\begin{align}
M_q &= D_{\theta_d}\big(F_{\theta_e}(I_q)\big), \\
M_q &\in \mathbb{R}^{H \times W \times N_c},
\end{align}
where $F_{\theta_e}$ denotes the encoder, $D_{\theta_d}$ the decoder, and $N_c$ the number of semantic classes. The segmentation model is treated as a fixed perceptual prior without task-specific architectural modifications.

\noindent \textbf{4D Cost Volume-based Pose Search.}
\sloppy
We sample the candidate pose $\boldsymbol{\xi}^{\text{samp}}$ around the pose prior $\boldsymbol{\xi}_{pri}$ to construct 4D cost volume $\mathcal{C}$. 
Within this search space, we deploy a 4D search algorithm. 
Namely, for each sampled camera pose $\boldsymbol{\xi}^{\text{samp}}_i$, we render the corresponding semantic mask $M_{r}^i$ from LoD city model and compare it with the segmentation mask of the input image $M_q$. 
The candidate pose achieving the highest similarity cost is selected as the coarse pose estimate $\boldsymbol{\xi}_c$. 
The pose vector $\boldsymbol{\xi}_{pri}$ is represented by a combination of translation and Euler angles:
\begin{equation}
\boldsymbol{\xi}_{pri} = (x_{pri}, y_{pri}, z_{pri}, \phi_{pri}, \theta_{pri}, \psi_{pri}),
\end{equation}
here, $(x_{pri}, y_{pri}, z_{pri})$ denotes the camera’s translation in 3D space, while $(\phi_{pri}, \theta_{pri}, \psi_{pri})$ correspond to the roll, pitch, and yaw angles, respectively. 
Since we keep $(\phi,\theta)$ fixed during search, uniform sampling is performed in the 4D space $[x, y, z, \psi]$ centered at the initial pose $\boldsymbol{\xi}_{pri}$. The sampling range is defined as
$
\mathbf{e}_c = \left[ e_{cx},\ e_{cy},\ e_{cz},\ e_{c \psi} \right]$,
and the number of samples in each dimension are 
$\mathbf{n}_c = \left[ n_{cx},\ n_{cy},\ n_{cz},\ n_{c \psi} \right]$,
where $\mathbf{e}_c$ denotes the sampling radius in each dimension, and $\mathbf{n}_c$ represents the corresponding number of samples. As a result, a $n_{cx} \times n_{cy} \times n_{cz} \times n_{c\psi}$ discrete 4D pose hypothesis set is generated, which will be used for subsequent cost volume construction and optimal pose selection. The $n_{cd}$ sampled poses along direction $d \in \{x, y, z, \psi\}$ is expressed as:
\begin{equation}
\left\{
\xi_{pri,d} - \frac{e_{cd}}{2},\ \cdots,\ 
\xi_{pri,d} + \frac{e_{cd}}{2}
\right\},
\label{eq:directional_sampling}
\end{equation}
For each sampled pose $\boldsymbol{\xi}_i^{\text{samp}} = (\mathbf{R}_i^{\text{samp}},\ \mathbf{t}_i^{\text{samp}})$, the rendered building mask $M_r(\boldsymbol{\xi}_i^{\text{samp}})$ and the semantic building mask $M_q$, the semantic cost $C_i^s$ is defined following IoU (Intersection over Union) as:
\begin{equation}
C_i^s =
\frac{1}{N_c}
\sum_{k=1}^{N_c}
\frac{
\left| M_r^k(\boldsymbol{\xi}_i^{\text{samp}}) \cap M_q^k \right|
}{
\left| M_r^k(\boldsymbol{\xi}_i^{\text{samp}}) \cup M_q^k \right|
}.
\label{eq:semantic_score}
\end{equation}
where $N_c$ denotes the number of semantic classes, $\cap$ denotes the pixel-wise intersection of the two binary masks, $\cup$ denotes the pixel-wise union of the two binary masks, and $\left| \cdot \right|$ denotes the number of foreground pixels in masks. 

Based on semantic cost of each sampled pose $\boldsymbol{\xi}_i^{\text{samp}}$, we construct a four-dimensional cost volume and search the coarse pose $\boldsymbol{\xi}_c$ with maximum value of semantic cost.
\subsection{Pose Refinement}
\label{Sec:refine}
\textbf{Joint Semantic–Depth Alignment.}
In the fine stage, we employ the model introduced in Section \ref{Sec:Coarse} to generate semantic masks at twice the spatial resolution.
The aforementioned segmentation mask (\cref{Sec:Coarse}) is extracted from the whole image, which struggles to capture local geometric features. To further improve the pose alignment accuracy, we introduce the depth matching leveraging the monocular depth estimator MoGe-2~\cite{wang2025moge}.

Specifically, we utilize MoGe-2 pretrained model to generate per-pixel depth map $D_q$ for the input UAV image $I_q$. The estimated depth values must be numerically aligned with the rendered depth map to ensure consistency in depth scale and range. For the depth map $D_r(\boldsymbol{\xi}_i^{\text{samp}})$ rendered at the candidate position $\boldsymbol{\xi}_i^{\text{samp}}$, and estimated depth map $D_q$, two global scale and shift parameters are estimated for each match, to calculate a robust depth cost:
\begin{equation}
C_i^d =
\sum_{p \in \mathcal{M}_{valid}}
\frac{1}{D_r(p)}
\left\|
s^* D_q(p) + t^* - D_r(p)
\right\|_1,
\end{equation}
where $\mathcal{M}_{valid}$ is the valid mask of rendered depth map, $D_r(p)$ denotes the rendered depth value at pixel $p$, and $s^*$ and $t^*$ are the optimal global scale and shift alignment estimated by least squares estimation for each match. The cost is inversely weighted by rendered depth to reduce the influence of distant, less reliable depth estimates.
The final alignment cost combines semantic and depth terms:
\begin{equation}
    C_i^{final} = \lambda_1 C_i^s + \lambda_2 C_i^d
\label{eq:final_cost}
\end{equation}
\noindent \textbf{Particle Filter Optimization.}
Inspired by other visual localization works~\cite{trivigno2024unreasonable,maggio2023loc,li2020hierarchical}, we design a particle filter based method for camera pose refinement. The core idea is to iteratively refine the initial hypothesis by introducing perturbations to the state, evaluating the resulting cost function, and updating the estimate toward convergence.
This stochastic refinement enables local exploration around the coarse hypothesis while avoiding exhaustive high-dimensional grid search, leading to stable and efficient convergence.

Specifically, having the coarse pose $\boldsymbol{\xi}_{c}$ selected from \cref{Sec:Coarse}, we introduce small perturbations $ \delta$ to its translation and rotation parameters, resulting in new candidate poses ${\boldsymbol{\xi}_{p_i}}$
\begin{equation}
    \left\{\boldsymbol{\xi}_{p_i}\right\}_{i=1}^{N_f}=\boldsymbol{\xi}_c+\left\{\delta \boldsymbol{\xi}_i\right\}_{i=1}^{N_f},
\end{equation}
where $\boldsymbol{\delta\xi}_i$ denotes the perturbation applied to the $i$-th particle in the pose space. 
Each perturbed pose $\boldsymbol{\xi}_{p_i}$ is then evaluated by rendering a semantic or geometric observation under that hypothesis and comparing it with the corresponding measurement from the query image using the predefined loss function derived from Eq.\eqref{eq:final_cost}. 
The camera pose is then updated in a weighted manner according to the probability scores obtained from the loss function. After performing $N_{iter}$ iterations, we obtain the final optimized pose estimate $\boldsymbol{\xi}^*$. 
\section{SemCityLockeD Benchmark Dataset}
\begin{figure}[htbp]
\centering
\begin{subfigure}[t]{0.240\textwidth}
    \includegraphics[width=\linewidth, keepaspectratio]{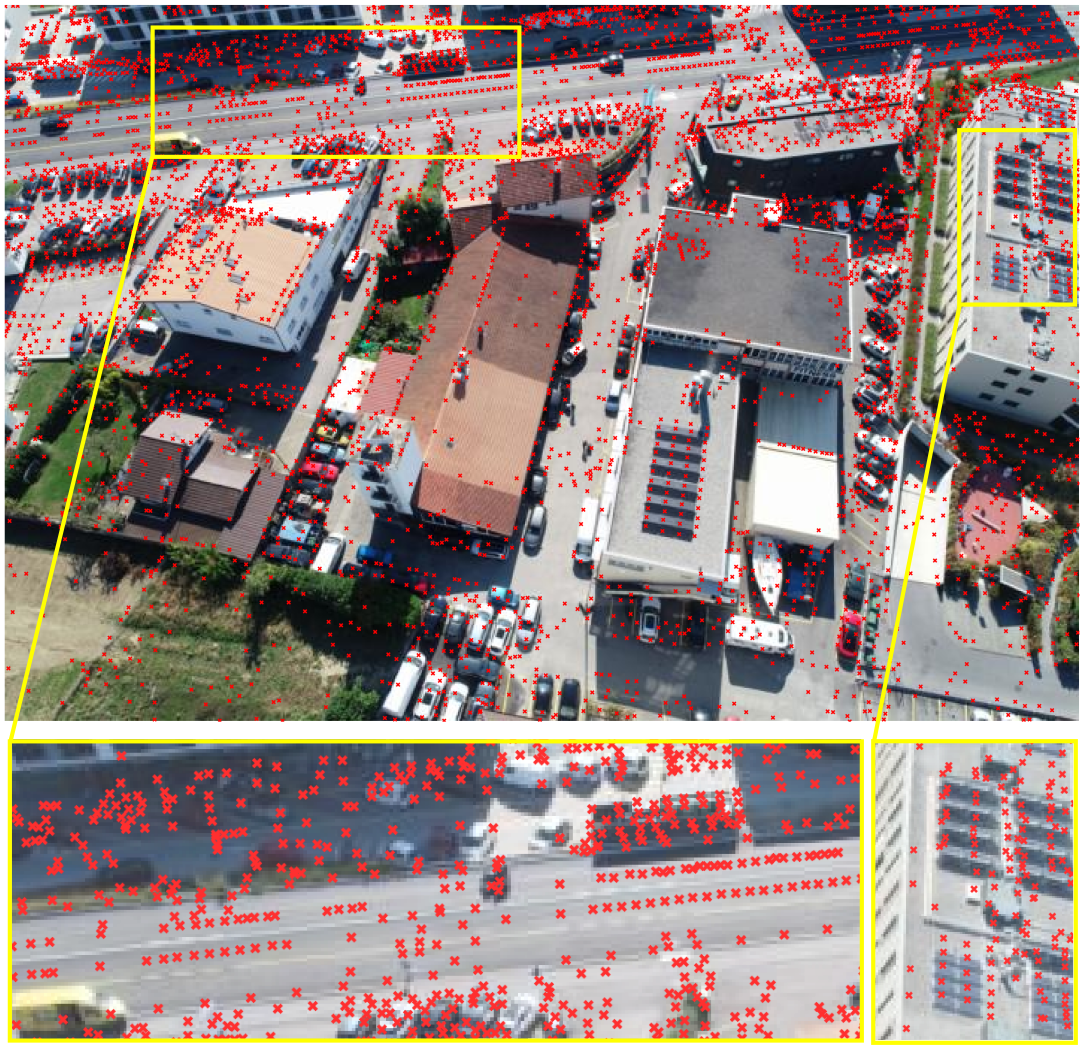}
    \caption{Swiss-EPFL~\cite{zhu2024lod,yan2022crossloc}}
\end{subfigure}
\hfill
\begin{subfigure}[t]{0.240\textwidth}
    \includegraphics[width=\linewidth, keepaspectratio]{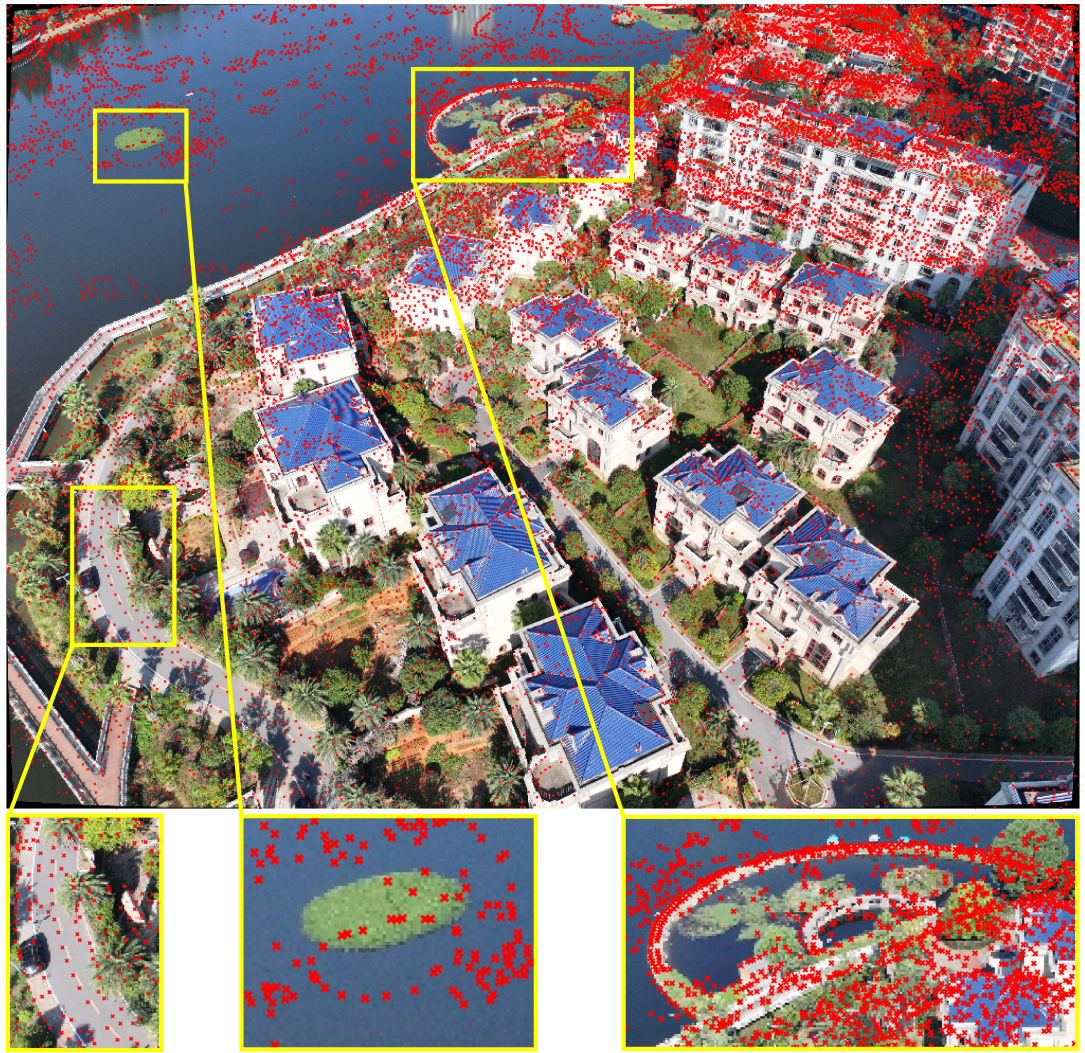}
    \caption{UAVD4L~\cite{zhu2024lod,wu2024uavd4l}}
\end{subfigure}
\hfill
\begin{subfigure}[t]{0.240\textwidth}
    \includegraphics[width=\linewidth, keepaspectratio]{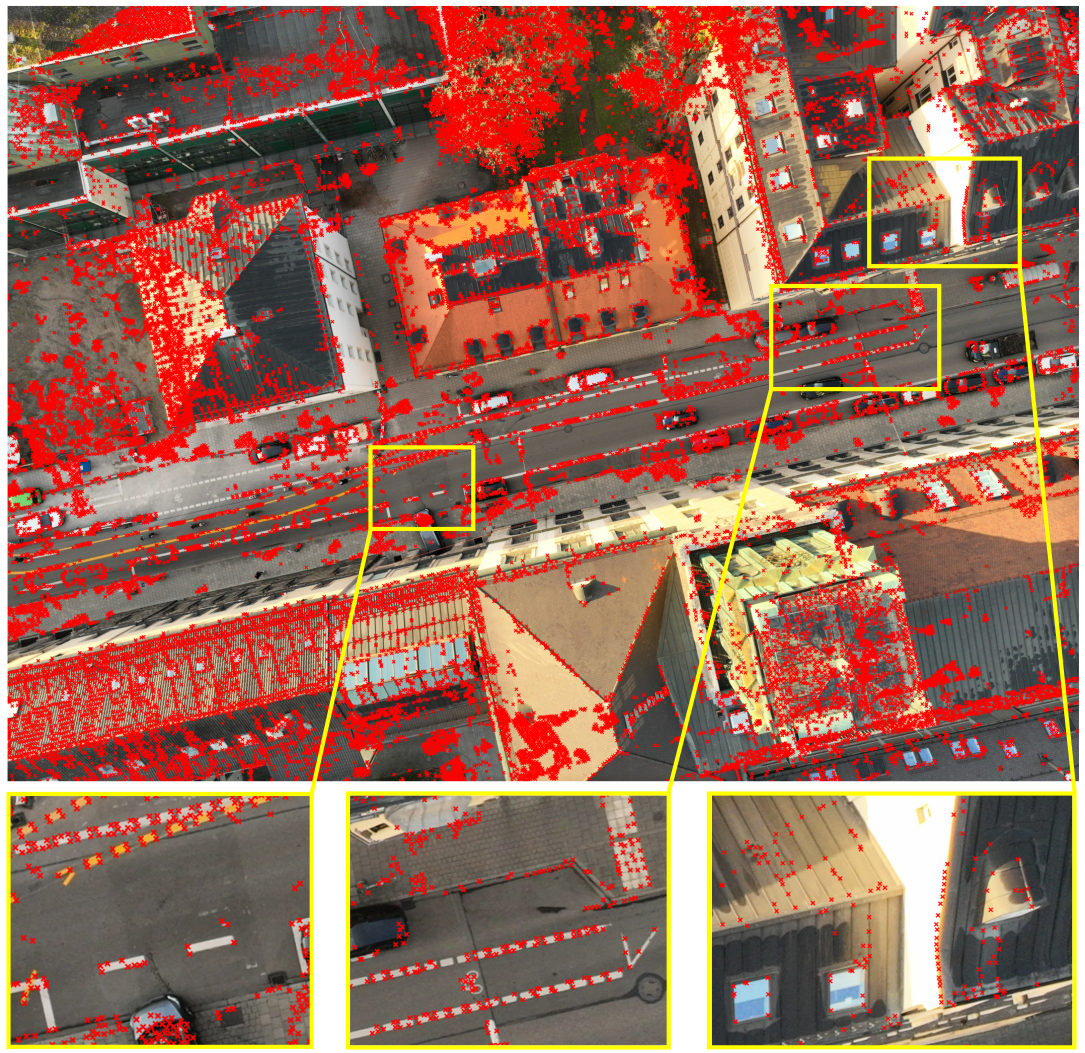}
    \caption{SemCityLockeD (nadir)}
\end{subfigure}
\hfill
\begin{subfigure}[t]{0.240\textwidth}
    \includegraphics[width=\linewidth, keepaspectratio]{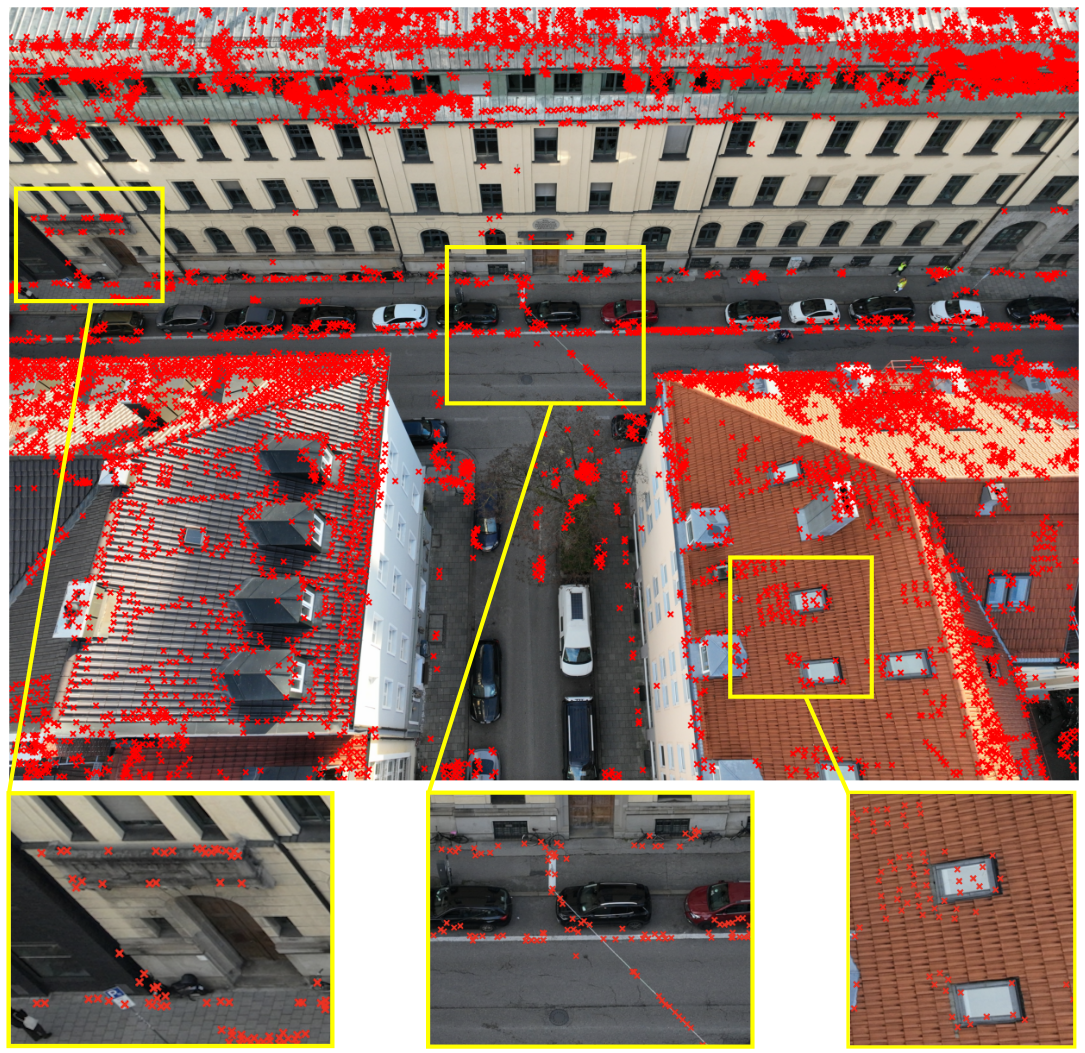}
    \caption{SemCityLockeD (oblique)}
\end{subfigure}
\caption{\textbf{Pose quality comparison.} Georeferenced keypoints are derived from corner detections on high-precision DOPs, lifted to 3D using DSMs, and then projected onto the UAV image using the ground-truth camera poses and intrinsics. In (c) and (d), corresponding to SemCityLockeD, the projected keypoints align closely with image structures, indicating high pose consistency. Note that (a) and (b) correspond to acquisitions at larger flight altitudes (above 75m) and in less dense urban settings, whereas (c) and (d) depict closer-range imagery in narrow urban canyon environments, representing substantially more challenging geometric conditions.}
\label{fig:accuracy_comparison_datasets}
\vspace{-1\baselineskip}
\end{figure}

We introduce \textbf{SemCityLockeD}, a novel real-world dataset to address the limitations of existing UAV localization benchmarks that lack accurate ground-truth poses, standardized semantic 3D city models, and realistic close-range urban canyon scenarios (\cref{fig:accuracy_comparison_datasets}).
SemCityLockeD captures a densely built urban environment characterized by closely spaced buildings, narrow urban canyons, and complex structural layouts spanning diverse architecture (19th-21st century) and building functions (residential, offices, shops, landmarks).
Data acquisition was performed using a DJI Matrice 350 RTK equipped with a Zenmuse L2 (4/3 CMOS) camera, flown at approximately 75\,m altitude in both nadir and oblique configurations. The dataset comprises 962 images (train/val/test: 586/188/188) with a resolution of $5{,}280 \times 3{,}956$ pixels, resulting in an average ground sampling distance (GSD) of 1.6\,cm. For privacy compliance, license plates and identifiable faces were removed from the imagery.
Camera poses were obtained via RTK-GNSS, inertial measurement unit (IMU) integration, and surveyed ground control points (GCPs), yielding georeferenced poses in EPSG:32632 with absolute accuracy of approximately 2\,cm. 
This high-accuracy georeferencing enables reliable pairing between UAV imagery and semantic 3D city models.

The corresponding LoD2 semantic 3D building models were acquired from the open data Bavarian state geoportal~\cite{LDBVlod2}, providing high absolute georeferenced footprint accuracy (approximately 2\,cm)~\cite{tum2twin,anders_uas_tum_downtown}. 
In addition to LoD2, we provide automatically generated LoD1 models and manually refined LoD3 models, enabling systematic evaluation across different LoDs and analysis of localization performance with respect to geometric abstraction, whereby both LoD1 and LoD3 were generated from LoD2 footprints to maintain the high accuracy and consistency.
Facade details were modeled based on additional street-level scanning point clouds which are not part of this dataset.

Beyond semantic LoD models, we further provide high-resolution textured mesh reconstructions generated from laser scanning and photogrammetric point clouds of the same UAV acquisition. 
These meshes complement the dataset allowing for rendering additional synthetic images, following the protocols of synthetic Swiss-EPFL~\cite{zhu2024lod,yan2022crossloc} and UAVD4L~\cite{zhu2024lod,wu2024uavd4l}.

Leveraging the precise pairing between UAV images and 3D city models, we additionally generate automatic semantic labels for the acquired imagery by projecting model semantics into image space. 
This enables large-scale, geometry-consistent annotation without manual labeling, facilitates supervised learning experiments, and provides a scalable annotation pipeline that can be readily adopted for future UAV datasets equipped with georeferenced 3D models.

In addition to the dataset, we will release a software suite for rendering semantic 3D city models into image and depth maps and pairing them with UAV observations, facilitating reproducibility and standardized evaluation. 
The full dataset and code will be made publicly available upon acceptance.
\section{Experimental Setup}
%
%
\noindent\textbf{Baselines.} We compare our proposed method with re-implemented visual localization baselines using semantic 3D building models with publicly available code: LoD-Loc~\cite{zhu2024lod}, MC-Loc~\cite{trivigno2024unreasonable}, and CAD-Loc~\cite{panek2023visual} (LoD-Loc v2 implementation unavailable at the time of writing~\cite{zhu2025lod}). 
All baseline methods, except for LoD-Loc, rely on feature-based extraction and matching approaches, in which the feature extraction algorithms include RoMa~\cite{edstedt2024roma} or e-LoFTR~\cite{wang2024efficient}.

\noindent\textbf{Data.} Besides evaluating on our introduced challenging scenario of the SemCityLockeD benchmark, we also evaluate on the openly available benchmarks that comprise 3D building models: UAVD4L-LoD and Swiss-EPFL~\cite{zhu2024lod} (the LoD-Loc v2 data is unavailable at the time of writing~\cite{zhu2025lod}).

\noindent\textbf{Metrics.} The evaluation metrics we follow are from the standard visual localization metrics~\cite{toft2020long} and we set the thresholds at $(2m,2^{\circ})$, $(3m,3^{\circ})$, $(5m,5^{\circ})$, same as the LoD-Loc baseline~\cite{zhu2024lod}. In addition, we also compute the average translation and yaw errors.
\section{Results and Discussion}

\noindent\textbf{Challenging Scenarios Performance.} 
\begin{figure}[h]
    \centering
    \includegraphics[width=1.0\linewidth]{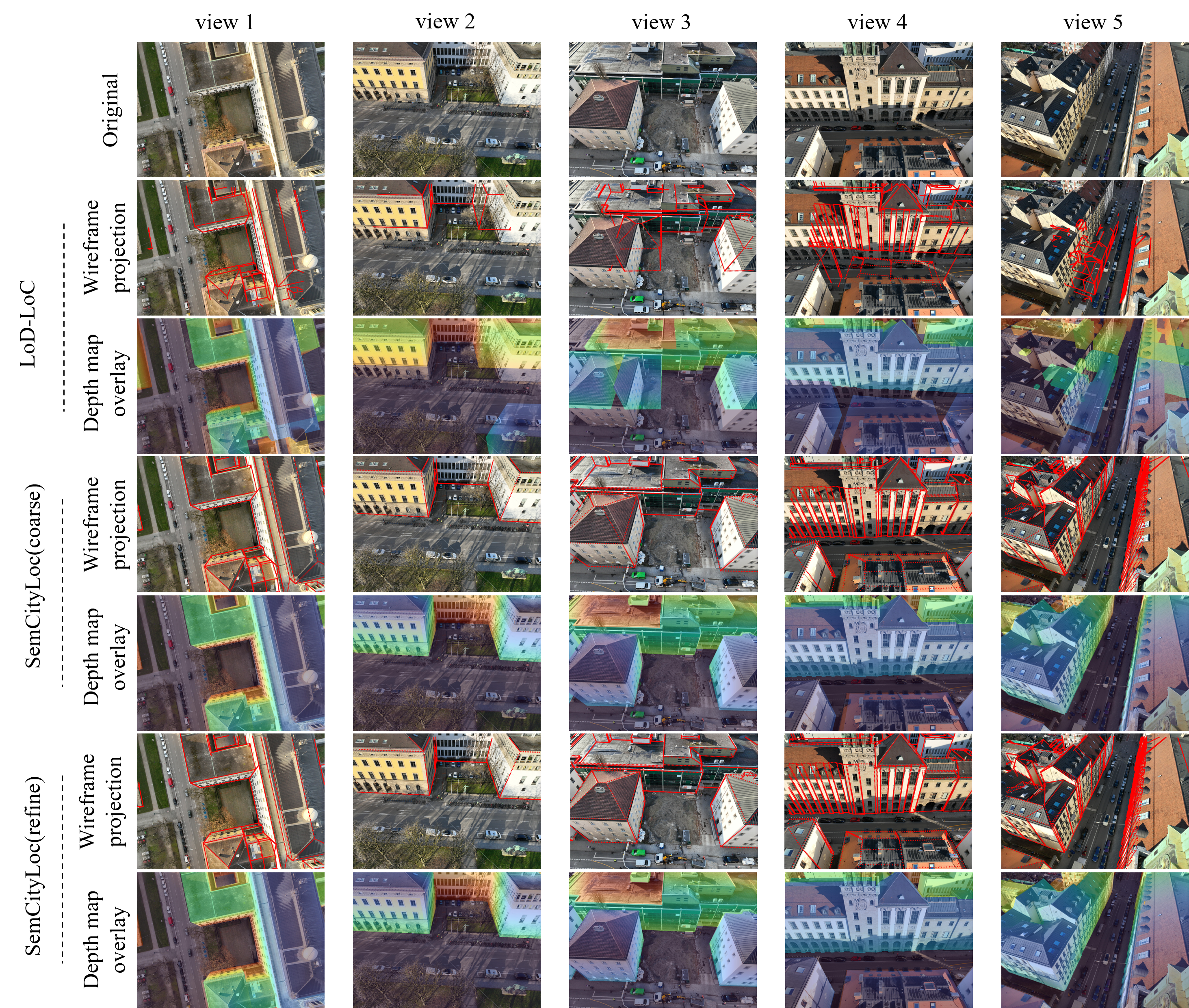}
    \caption{\textbf{Qualitative comparison between LoD-Loc~\cite{zhu2024lod} and SemCityLoc.}
Projected LoD geometry using the estimated poses reveals noticeable misalignment for LoD-Loc, especially in oblique views and facade-occluded urban canyon scenes. 
In contrast, SemCityLoc produces geometrically consistent overlays and accurate depth alignment, demonstrating improved robustness to perspective distortion and structural repetition (especially apparent in view 3).} 
    \label{fig:visualization_results}
    \vspace{-0.7\baselineskip}
\end{figure}
We evaluate SemCityLoc on the proposed SemCityLockeD benchmark, which represents densely built urban canyon environments with strong geometric ambiguities. 
As shown in \Cref{tab:SemCity_result}, our method consistently outperforms all baselines across all recall thresholds and error metrics.
At the strict $2$m–$2^{\circ}$ threshold, SemCityLoc achieves 69.15\% recall compared to 35.11\% for LoD-Loc~\cite{zhu2024lod}, nearly doubling performance. 
Recall further improves to 84.04\% and 89.36\% at $3$m–$3^{\circ}$ and $5$m–$5^{\circ}$, respectively. 
Mean yaw error decreases from 1.78$^\circ$ to 0.42$^\circ$, and translation error from 9.89\,m to 2.62\,m.
These results demonstrate that integrating semantic segmentation and monocular depth provides substantially stronger geometric constraints than wireframe-based alignment alone, particularly in close-range urban scenes with facade repetitions and limited visibility.
\begin{table*}[htbp]
\centering

\caption{\textbf{Camera pose evaluation on the SemCityLockeD dataset.} Pose recall is reported in \%, rotation error (yaw) in degrees, and positional errors in meters. Our SemCityLoc significantly outperforms all baselines across all metrics. \textbf{Bold} = best result; \underline{underlined} = second best.}
\label{tab:SemCity_result}
\scriptsize
\resizebox{0.7\textwidth}{!}{%
\begin{tabular}{ll|ccc|cc}
\toprule
\textbf{Method} & & \multicolumn{3}{c|}{\textbf{Recall (\%)}} & \multicolumn{2}{c}{\textbf{Mean Error}} \\
\cmidrule(lr){3-5} \cmidrule(lr){6-7}
\textbf{Metrics}& & \textbf{2m-2°} & \textbf{3m-3°} & \textbf{5m-5°} & \textbf{Yaw (°)} & \textbf{XYZ (m)} \\
\midrule
\multirow{2}{*}{\makecell{\textbf{CAD-Loc}\\ \textit{LoD model}}}
    & e-LoFTR      & 0  & 0  & 0  & -- & -- \\
    & RoMa         & 0  & 0  & 0  & -- & -- \\
\midrule
\multirow{2}{*}{\makecell{\textbf{MC-Loc}\\ \textit{LoD model}}}
    & DINOv2       & 5.32  & 8.51  & 18.09 & 5.90 & 10.71  \\
    & RoMa         & 0  & 1.06  & 3.19  & 6.67 & 18.61  \\
\midrule
\makecell{\textbf{LoD-Loc}\\ \textit{LoD model}}
    & LoD-Loc      & 35.11 & 47.87 & 53.19 & 1.78 & 9.89 \\
\midrule
\multirow{3}{*}{\makecell{\textbf{SemCityLoc}\\ \textit{LoD model}}}

    & \cellcolor[gray]{0.95}\textit{No Coarse Selection} 
    & \cellcolor[gray]{0.95}41.49 
    & \cellcolor[gray]{0.95}56.38  
    & \cellcolor[gray]{0.95}72.34 
    & \cellcolor[gray]{0.95}1.11  
    & \cellcolor[gray]{0.95}5.94  
    \\

    & \cellcolor[gray]{0.95}\textit{No Refinement}     
    & \cellcolor[gray]{0.95}\underline{56.38} 
    & \cellcolor[gray]{0.95}\underline{70.21} 
    & \cellcolor[gray]{0.95}\underline{81.91} 
    & \cellcolor[gray]{0.95}\underline{1.04}  
    & \cellcolor[gray]{0.95}\underline{3.20}  
    \\

    & \cellcolor[gray]{0.95}\textit{Full model}    
    & \cellcolor[gray]{0.95}\textbf{69.15}  
    & \cellcolor[gray]{0.95}\textbf{84.04} 
    & \cellcolor[gray]{0.95}\textbf{89.36} 
    & \cellcolor[gray]{0.95}\textbf{0.42}  
    & \cellcolor[gray]{0.95}\textbf{2.62}  
    \\
\bottomrule
\end{tabular}}
\end{table*}
Ablation results confirm the importance of both modules (\cref{tab:SemCity_result,fig:visualization_results}). 
Coarse semantic selection increases recall at $2$m–$2^{\circ}$ from 41.49\% to 56.38\%, while semantic–depth refinement further improves it to 69.15\%. 
Both stages are therefore complementary and jointly critical for stable pose estimation.

Qualitative results in \Cref{fig:visualization_results} further corroborate the improvements. In particular, View~3 highlights a challenging oblique urban canyon configuration with strong perspective distortion and facade self-occlusion, where the baseline fails to produce a consistent geometric alignment. 
In contrast, SemCityLoc achieves precise overlay between projected LoD geometry and image structures, demonstrating robust semantic–geometric alignment even under severe oblique viewing conditions and repetitive facade patterns.

\noindent\textbf{Efficiency.} 
SemCityLoc achieves markedly improved training efficiency compared with LoD-Loc~\cite{zhu2024lod}. 
\begin{figure}[htb]
    \centering
    \includegraphics[width=0.6\linewidth]{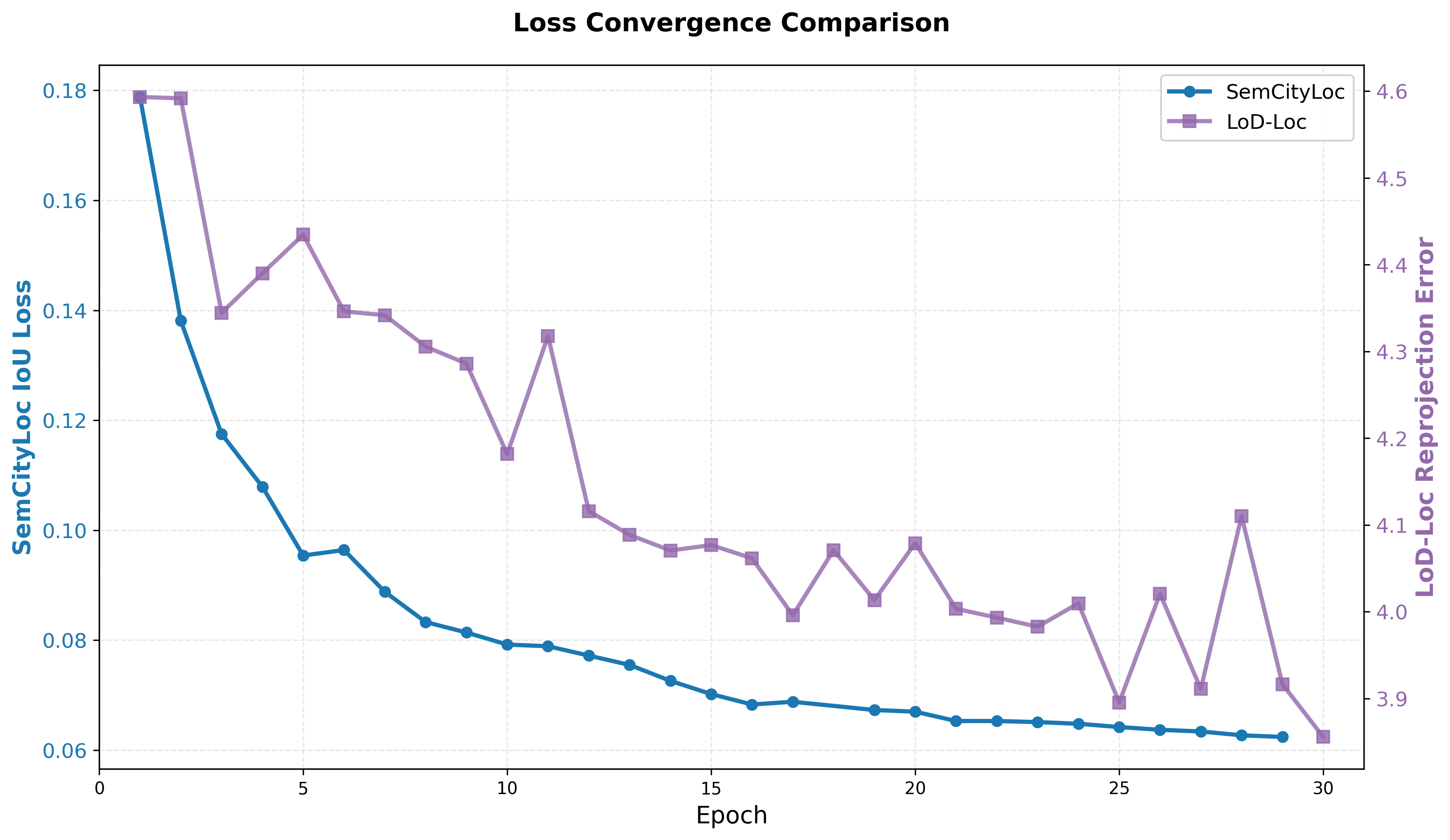}
    \caption{\textbf{Training convergence on SemCityLockeD.} 
SemCityLoc exhibits faster and more stable optimization compared to LoD-Loc, reflecting the benefit of fine-tuning a lightweight head on pretrained features with IoU-based alignment.}
    \label{fig:covergence}
    \vspace{-0.8\baselineskip}
\end{figure}
As shown in \Cref{fig:covergence}, SemCityLoc converges within the first few epochs and stabilizes after approximately 15 epochs, whereas LoD-Loc exhibits slower convergence and pronounced loss oscillations. 
This behavior can be attributed to two factors.
First, SemCityLoc optimizes only a lightweight segmentation head on top of pretrained foundation models, effectively operating in a well-conditioned representation space. 
In contrast, LoD-Loc is trained from scratch, requiring simultaneous feature learning and geometric alignment, which increases optimization complexity and susceptibility to local minima.
Second, our IoU-based semantic alignment objective induces smoother and more globally structured gradients than reprojection-based losses. 
Semantic masks provide dense, region-level supervision that is less sensitive to small geometric perturbations, whereas reprojection losses rely on sparse edge correspondences and exhibit higher non-convexity with respect to pose parameters. 
As a result, IoU-driven optimization yields more stable gradient behavior and faster convergence.

\noindent\textbf{Generalization to Sub-Urban and High-Altitude Scenarios.}
We further evaluate SemCityLoc on the UAVD4L-LoD and Swiss-EPFL datasets comprising mixed urban scenarios to assess generalization beyond challenging close-range urban canyons. 
As reported in~\cite{zhu2024lod}, LoD-Loc achieves strong performance on these benchmarks and is therefore adopted as our primary reference baseline.

As shown in \Cref{tab:UAVDL_result}, SemCityLoc achieves competitive or superior performance compared to LoD-Loc in both in-Traj and out-of-Traj settings on UAVD4L-LoD. In the in-Traj scenario, our method improves most recall thresholds and reduces translation error, while maintaining comparable rotation accuracy. In the out-of-Traj setting, SemCityLoc outperforms the baseline on nearly all metrics, demonstrating improved robustness to trajectory shifts.
Results on the Swiss-EPFL dataset (\cref{tab:Swiss_result}) further confirm the benefits of integrating semantics and depth cues. In the in-Place scenario, recall values are comparable to LoD-Loc, while translation and heading errors are consistently reduced. In the challenging out-of-Place setting, SemCityLoc significantly improves recall (up to two-fold) and substantially reduces mean positional error from 12.54\,m to 3.09\,m (\cref{tab:Swiss_result}).

Overall, these results indicate that the proposed semantic–geometric alignment strategy generalizes effectively across heterogeneous environments comprising mixed urban and sub-urban regions, varying structural densities, and diverse acquisition altitudes, while consistently outperforming wireframe-based localization methods.

We note that UAVD4L-LoD and Swiss-EPFL provide lower-precision ground-truth poses compared to SemCityLockeD, which may introduce additional evaluation noise. 
Nevertheless, consistent improvements across datasets demonstrate the robustness of our approach. Additional details are provided in Suppl.
\begin{table*}[htp]
\centering
\caption{\textbf{Camera pose evaluation on the UAVD4L-LoD dataset.} Pose recall is reported in \%, rotation error (yaw) in degrees, and positional errors in meters. Our SemCityLoc achieves competitive performance, outperforming the baseline in most metrics. \textbf{Bold} = best result; \underline{underlined} = second best.}
\label{tab:UAVDL_result}
\scriptsize
\resizebox{0.90\textwidth}{!}{%
\begin{tabular}{ll|ccc|cc|ccc|cc}
\toprule
\multicolumn{2}{l|}{\textbf{Method}}& \multicolumn{5}{c|}{\textbf{in-Traj}} & \multicolumn{5}{c}{\textbf{out-of-Traj}} \\
\cmidrule(lr){3-7} \cmidrule(lr){8-12}
& &  \multicolumn{3}{c|}{\textbf{Recall (\%)}} & \multicolumn{2}{c|}{\textbf{Mean Error}} &  \multicolumn{3}{c|}{\textbf{Recall (\%)}} & \multicolumn{2}{c}{\textbf{Mean Error}}\\
\cmidrule(lr){3-5} \cmidrule(lr){6-7} \cmidrule(lr){8-10} \cmidrule(lr){11-12}
\textbf{Metrics}& & \textbf{2m-2°} & \textbf{3m-3°} & \textbf{5m-5°} &\textbf{Yaw (°)} & \textbf{XYZ (m)}& \textbf{2m-2°} & \textbf{3m-3°} & \textbf{5m-5°} & \textbf{Yaw (°)} & \textbf{XYZ (m)} \\
\midrule
{\makecell{\textbf{LoD-Loc} \\ \textit{LoD model}}}
    & LoD-Loc       & \textbf{83.42} & \underline{90.65} & 96.57 & \underline{0.19}  &\underline{1.73} &\underline{88.18} & \textbf{98.49} & \underline{99.54} & \underline{0.18} & \underline{1.38}  \\
\midrule
\multirow{3}{*}{\makecell{\textbf{SemCityLoc} \\ \textit{LoD model}}}
    & \cellcolor[gray]{0.95}\textit{No Coarse Selection} & \cellcolor[gray]{0.95}62.03 & \cellcolor[gray]{0.95}89.96 & \cellcolor[gray]{0.95}\underline{99.05} & \cellcolor[gray]{0.95}0.20& \cellcolor[gray]{0.95}1.84 &\cellcolor[gray]{0.95} 47.63 &\cellcolor[gray]{0.95} 62.32 &\cellcolor[gray]{0.95} 78.51 &\cellcolor[gray]{0.95} 0.34 &\cellcolor[gray]{0.95} 3.36 \\
    & \cellcolor[gray]{0.95}\textit{No Refinement}     &\cellcolor[gray]{0.95} 47.07 &\cellcolor[gray]{0.95} 74.88 &\cellcolor[gray]{0.95} 91.21 & \cellcolor[gray]{0.95}0.34  &\cellcolor[gray]{0.95} 2.48 &\cellcolor[gray]{0.95} 48.49 &\cellcolor[gray]{0.95} 83.94 &\cellcolor[gray]{0.95} 99.13 &\cellcolor[gray]{0.95} 0.68  & \cellcolor[gray]{0.95}2.18 \\
    &\cellcolor[gray]{0.95} \textit{Full model}  &\cellcolor[gray]{0.95}\underline{70.45} &\cellcolor[gray]{0.95} \textbf{92.96} &\cellcolor[gray]{0.95} \textbf{99.19}&\cellcolor[gray]{0.95} \textbf{0.19}  &\cellcolor[gray]{0.95}\textbf{1.68}  &\cellcolor[gray]{0.95} \textbf{89.64}  &\cellcolor[gray]{0.95} \underline{98.40} &\cellcolor[gray]{0.95} \textbf{99.95} &\cellcolor[gray]{0.95}\textbf{0.14} &\cellcolor[gray]{0.95}\textbf{1.26} \\
\bottomrule
\end{tabular}
}
\end{table*}
\begin{table*}[htp]
\centering
\caption{\textbf{Camera pose evaluation on the Swiss-EPFL dataset.} Pose recall is reported in \%, rotation error (yaw) in degrees, and positional errors in meters. Our SemCityLoc outperforms the baseline in most metrics, with particularly strong performance on out-of-place scenarios. \textbf{Bold} = best result; \underline{underlined} = second best.}
\label{tab:Swiss_result}
\scriptsize
\resizebox{0.90\textwidth}{!}{%
\begin{tabular}{ll|ccc|cc|ccc|cc}
\toprule
\multicolumn{2}{l|}{\textbf{Method}}& \multicolumn{5}{c|}{\textbf{in-Place}} & \multicolumn{5}{c}{\textbf{out-of-Place}} \\
\cmidrule(lr){3-7} \cmidrule(lr){8-12}
& &  \multicolumn{3}{c|}{\textbf{Recall (\%)}} & \multicolumn{2}{c|}{\textbf{Mean Error}} &  \multicolumn{3}{c|}{\textbf{Recall (\%)}} & \multicolumn{2}{c}{\textbf{Mean Error}}\\
\cmidrule(lr){3-5} \cmidrule(lr){6-7} \cmidrule(lr){8-10} \cmidrule(lr){11-12}
\textbf{Metrics}& & \textbf{2m-2°} & \textbf{3m-3°} & \textbf{5m-5°} &\textbf{Yaw (°)} & \textbf{XYZ (m)}& \textbf{2m-2°} & \textbf{3m-3°} & \textbf{5m-5°} & \textbf{Yaw (°)} & \textbf{XYZ (m)} \\
\midrule
\makecell{\textbf{LoD-Loc}\\\textit{LoD model}}
& LoD-Loc & \textbf{36.23} & 50.00 & \textbf{63.62} & 3.73 & \underline{9.17} & \underline{17.41} & \underline{30.61} & \underline{48.55} & 3.40 & 12.54 \\
\midrule

\multirow{3}{*}{\makecell{\textbf{SemCityLoc}\\\textit{LoD model}}}
& \cellcolor[gray]{0.95}\textit{No Coarse Selection} 
& \cellcolor[gray]{0.95}5.34  
& \cellcolor[gray]{0.95}15.73 
& \cellcolor[gray]{0.95}31.32 
& \cellcolor[gray]{0.95}2.15  
& \cellcolor[gray]{0.95}11.60 
& \cellcolor[gray]{0.95}6.86 & \cellcolor[gray]{0.95}17.15 & \cellcolor[gray]{0.95}35.88 & \cellcolor[gray]{0.95}1.62 & \cellcolor[gray]{0.95}9.05 \\

& \cellcolor[gray]{0.95}\textit{No Refinement} 
& \cellcolor[gray]{0.95}10.11 
& \cellcolor[gray]{0.95}20.51  
&\cellcolor[gray]{0.95}35.96 
& \cellcolor[gray]{0.95}\underline{2.04} 
& \cellcolor[gray]{0.95}10.39
& \cellcolor[gray]{0.95}8.71 & \cellcolor[gray]{0.95}19.79 & \cellcolor[gray]{0.95}42.22 & \cellcolor[gray]{0.95}\underline{1.23} & \cellcolor[gray]{0.95}\underline{7.86} \\

& \cellcolor[gray]{0.95}\textit{Full model} 
& \cellcolor[gray]{0.95}\underline{33.99} & \cellcolor[gray]{0.95}\textbf{50.70} & \cellcolor[gray]{0.95}\underline{62.36} 
& \cellcolor[gray]{0.95}\textbf{1.31} & \cellcolor[gray]{0.95}\textbf{7.13} 
& \cellcolor[gray]{0.95}\textbf{35.36} & \cellcolor[gray]{0.95}\textbf{62.01} & \cellcolor[gray]{0.95}\textbf{89.18} 
& \cellcolor[gray]{0.95}\textbf{0.93} & \cellcolor[gray]{0.95}\textbf{3.09} \\

\bottomrule
\end{tabular}
}
\end{table*}

\noindent\textbf{Sensitivity Analysis to Segmentation Quality.}
We evaluate the impact of semantic quality by training a lightweight head on top of a shared DINOv3 backbone per dataset, achieving 88\% mIoU on SemCityLockeD, 85\% on UAVD4L-LoD, and 78\% on Swiss-EPFL—sufficient for stable semantic–geometric alignment.

Fully zero-shot alternatives perform substantially worse: CLIP \cite{radford2021learningCLIP} + Semantic-SAM \cite{li2024segmentSemSam} yields 32\% mIoU on SemCityLockeD and reduces recall to 15.93\% / 28.57\% / 52.74\% at 2m–2° / 3m–3° / 5m–5°, while Grounded-SAM2 \cite{ren2024grounded} achieves 37\% mIoU with unstable masks on the same dataset. We attribute this degradation to domain mismatch between terrestrial pretraining data and UAV viewpoints. These results indicate that moderate dataset-adaptive supervision suffices for reliable localization, whereas fully zero-shot semantics remain insufficient (more details in Suppl.).

\noindent\textbf{Sensitivity Analysis to Pose Prior Noise.}
The translation priors are initially perturbed by 5–15 m, on top of which we inject varying independent perturbation of up to 200 m per axis. Rotations are also perturbed by 0–8°(\cref{tab:pose_prior_all}).
\begin{table}[t]
\centering
\caption{\textbf{Pose prior sensitivity analysis.} Localization recall (\%) under increasing translational perturbations (up to 200\,m per axis).}
\label{tab:pose_prior_all}
\setlength{\tabcolsep}{4pt}
\scriptsize
\resizebox{0.9\textwidth}{!}{
\begin{tabular}{l l ccccccc}
\toprule
\textbf{Metric}& & \multicolumn{7}{c}{\textbf{Perturbation Range per Axis (m)}} \\ 
\cmidrule(lr){3-9}
\textbf{Threshold}($\Delta$m)&  & \textbf{10--20} & \textbf{20--30} & \textbf{30--40} & \textbf{40--50} & \textbf{50--100} & \textbf{100--150} & \textbf{150--200} \\
\midrule
\multirow{3}{*}{\textbf{Recall (\%)}} 
& 2m--2$^\circ$ & \cellcolor[gray]{0.95}68.09 & \cellcolor[gray]{0.95}67.02 & \cellcolor[gray]{0.95}65.96 & \cellcolor[gray]{0.95}62.77 & \cellcolor[gray]{0.95}45.74 & \cellcolor[gray]{0.95}29.79 & \cellcolor[gray]{0.95}18.09 \\

& 3m--3$^\circ$ & \cellcolor[gray]{0.95}80.85 & \cellcolor[gray]{0.95}75.53 & \cellcolor[gray]{0.95}73.40 & \cellcolor[gray]{0.95}70.21 & \cellcolor[gray]{0.95}50.00 & \cellcolor[gray]{0.95}30.85 & \cellcolor[gray]{0.95}18.09 \\

& 5m--5$^\circ$ & \cellcolor[gray]{0.95}87.23 & \cellcolor[gray]{0.95}81.91 & \cellcolor[gray]{0.95}78.72 & \cellcolor[gray]{0.95}76.60 & \cellcolor[gray]{0.95}53.19 & \cellcolor[gray]{0.95}36.17 & \cellcolor[gray]{0.95}19.15 \\
\bottomrule
\end{tabular}
}
\end{table}
The pose prior sensitivity analysis shows that recall degrades smoothly as translational perturbations increase. Even with perturbations up to 50 m per axis, the method maintains relatively stable performance, with recall remaining at 62.77\%, 70.21\%, and 76.60\% under the 2 m--2°, 3 m--3°, and 5 m--5° thresholds, respectively. This demonstrates robustness to substantial initialization errors. However, when the pose prior error exceeds 100 m, localization recall drops dramatically across all thresholds. This sharp decrease indicates that very large prior errors significantly affect localization reliability, revealing an operational boundary of the method.

\noindent\textbf{Sensitivity Analysis Across LoDs.}
We analyze how structural abstraction in the 3D city model affects pose alignment (\cref{tab:LoDs}). While LoD2 offers a favorable trade-off between availability and fidelity, SemCityLockeD uniquely provides standardized LoD1–LoD3 models~\cite{wysocki2024reviewing}, enabling systematic evaluation.
\begin{table}[htbp]
\centering
\caption{\textbf{Localization performance across different LoDs.} Increasing geometric detail strengthens pose constraints, while semantic LoD2 achieves performance comparable to LoD3 despite reduced geometric complexity.}
\label{tab:LoDs}
\scriptsize
\resizebox{0.5\textwidth}{!}{%
\begin{tabular}{l|ccc|cc}
\toprule
\textbf{LoD Model} & \multicolumn{3}{c|}{\textbf{Recall (\%)}} & \multicolumn{2}{c}{\textbf{Mean Error}} \\
\cmidrule(lr){2-4} \cmidrule(lr){5-6}
\textbf{Metrics} & \textbf{2m-2°} & \textbf{3m-3°} & \textbf{5m-5°} & \textbf{Yaw (°)} & \textbf{XYZ (m)} \\
\midrule
\cellcolor[gray]{0.95}\textit{LoD1} 
& \cellcolor[gray]{0.95}48.94 
& \cellcolor[gray]{0.95}54.26
& \cellcolor[gray]{0.95}69.15
& \cellcolor[gray]{0.95}1.14
& \cellcolor[gray]{0.95}6.01 \\

\cellcolor[gray]{0.95}\textit{LoD3}     
& \cellcolor[gray]{0.95}\textbf{70.21}
& \cellcolor[gray]{0.95}\underline{75.53}
& \cellcolor[gray]{0.95}\underline{89.36}
& \cellcolor[gray]{0.95}\underline{0.50}
& \cellcolor[gray]{0.95}\textbf{2.48} \\

\cellcolor[gray]{0.95}\textit{LoD2(semantic)}    
& \cellcolor[gray]{0.95}\underline{69.15}
& \cellcolor[gray]{0.95}\textbf{84.04}
& \cellcolor[gray]{0.95}\textbf{89.36}
& \cellcolor[gray]{0.95}\textbf{0.42}
& \cellcolor[gray]{0.95}\underline{2.62} \\
\bottomrule
\end{tabular}}
\end{table}
Performance improves from LoD1 to LoD3, reflecting increased structural discriminability. Notably, semantic LoD2 achieves accuracy close to or even surpasses that of LoD3 despite reduced geometric resolution. This indicates that semantically partitioned surface structure enhances pose observability and can partially compensate for missing fine-grained geometry. Pose accuracy is therefore governed not only by geometric density, but by the joint availability of discriminative structural cues and semantically coherent surface representations (more details in Suppl.).

\noindent\textbf{Running time analysis.}
Using nvdiffrast-based~\cite{laine2020modular} GPU rasterization and frustum culling, SemCityLoc requires only 0.878\,s per image, including 0.116\,s for model inference, 0.409\,s for coarse pose search, and 0.353\,s for particle-filter refinement. The resulting sub-second runtime highlights the feasibility of deploying semantic city-model-based localization in practical UAV systems, where accurate relocalization must be performed online under large pose uncertainty.

\noindent\textbf{Limitations and Future Work.}
%
%
SemCityLoc relies on the availability and geometric accuracy of semantic 3D city models, and variations in model completeness or semantic richness may affect localization robustness across regions. While the method is generally robust to dynamic objects absent from the LoD models (e.g., vehicles and vegetation), performance may degrade in rare cases with severely limited geometric observability, such as scenes containing only a single visible building or facade. Finally, localization accuracy depends on the quality of semantic segmentation and depth estimation, although continued advances in foundation models are expected to further improve robustness.
\section{Conclusion}
We present SemCityLoc, a lightweight semantic–geometric localization system that leverages standardized 3D city models for accurate and scalable aerial 6DoF pose estimation. 
By jointly aligning semantic segmentation and monocular depth with structured LoD representations, our method enables robust localization without dense radiometric reconstructions.
We further introduce SemCityLockeD, a unique real-world benchmark comprising centimeter-accurate UAV poses, close-range urban canyon imagery, and standardized LoD1–LoD3 models. 
Across challenging scenarios, SemCityLoc improves recall at 2m–2° from 35.11\% to 69.15\%, demonstrating substantially stronger pose discriminability than existing map-based approaches while remaining robust to segmentation noise and prior inaccuracies.
Overall, our results indicate that semantically structured 3D city models provide sufficient and scalable constraints for high-precision aerial localization, establishing structured semantic–geometric alignment as a promising direction for next-generation map-based aerial localization systems.

\section*{Acknowledgements}
This work was supported in part by the Cambridge Centre for Smart Infrastructure and Construction (CSIC), the Laing O’Rourke Centre for Construction Engineering and Technology at the University of Cambridge; the Anhui Provincial Natural Science Foundation (Grant No. 2508085MF142), and the USTC Global Initiative Fund (MS-A-2026-1-19) at the University of Science and Technology of China. The authors thank the members of the Computer Vision for Digital Twins (CV4DT) group at the University of Cambridge for their support and valuable discussions. The authors also thank Yuhan Jiang for assistance with figure preparation. The authors further acknowledge the contributors to the TUM2TWIN project \url{https://tum2t.win/}, in particular Katharina Anders, Jiapan Wang, Xiaoyu Huang, and Sidi Liu, for their efforts in processing and publishing the data \url{https://zenodo.org/records/14548134}.

\bibliographystyle{splncs04}
\bibliography{main}
\end{document}